\pdfoutput=1

\documentclass[11pt]{article}

\usepackage{ACL2023}

\usepackage{times}
\usepackage{latexsym}
\usepackage{graphicx}
\usepackage{booktabs}
\usepackage[T1]{fontenc}

\usepackage[utf8]{inputenc}

\usepackage{microtype}

\usepackage{inconsolata}

\title{Evaluating the Representation of Vowels in Wav2Vec Feature Extractor: A Layer-Wise Analysis Using MFCCs}

\author{Domenico De Cristofaro \\
  Free University of Bozen  \\\And
  Vincenzo Norman Vitale  \\
  University of Naples Federico II  \\\And
  Alessandro Vietti  \\
  Free University of Bozen \\}

\begin{document}
\maketitle
\begin{abstract}
Automatic Speech Recognition has advanced with self-supervised learning, enabling feature extraction directly from raw audio. In Wav2Vec, a CNN first transforms audio into feature vectors before the transformer processes them. This study examines CNN-extracted information for monophthong vowels using the TIMIT corpus. We compare MFCCs, MFCCs with formants, and CNN activations by training SVM classifiers for front-back vowel identification, assessing their classification accuracy to evaluate phonetic representation.
\end{abstract}

\section{Introduction}

Self-supervised ASR models like Wav2Vec, HuBERT, and WavLM learn speech representations directly from raw audio, surpassing the need for handcrafted features. In contrast, hybrid ASR systems relied on MFCCs \cite{Povey_ASRU2011, Chatterjee2009}, which some models, like HuBERT, still incorporate for their phonetic relevance. While transformer layers in self-supervised models are well-studied, the role of convolutional feature extractors remains underexplored, despite their crucial function in shaping speech representations for the transformer layers. Wav2Vec’s feature encoder consists of seven convolutional layers that hierarchically transform raw waveforms into feature vectors before contextual modeling in transformers. Understanding how these layers encode phonetic information is essential for ASR development. This study investigates the relationship between low-level acoustic features (MFCCs) and CNN activations in Wav2Vec. Using SVM classifiers, we compare MFCCs, MFCCs with formants (F1, F2), and CNN outputs to assess their effectiveness in front-back vowel classification.
\section{Related Work}

Recent studies have used probing techniques to investigate how transformer-based ASR models encode phonetic information \cite{belinkov2017, pasad2022layerwiseanalysisselfsupervisedspeech, scharenborg2019representation}. Early work \cite{belinkov2017} revealed differences across layers in phoneme category prediction, while later studies examined phonetic and graphemic representations \cite{belinkov2020analyzingphoneticgraphemicrepresentations} and consonant-vowel distinctions using PCA-transformed embeddings \cite{scharenborg2019representation}. While prior research has focused on transformers, convolutional feature extractors in self-supervised ASR remain underexplored. Unlike past work, this study analyzes intermediate CNN activations in Wav2Vec, comparing them with MFCCs and formants to evaluate their role in phonetic discrimination. Recent analyses show that self-supervised convolutional layers capture key spectral features, with intermediate layers demonstrating higher correlations \cite{vieting2023comparativeanalysiswav2vec20}. \cite{english2022domain} found that transformers encode phonetic details for manner and place of articulation, while \cite{english2024searching} applied association rule mining to reveal linguistic structures. Other works have examined transformer layers' ability to detect hesitation \cite{vitale24_interspeech} and syllables \cite{vitale2024exploring}, yet the feature extractor remains largely unexamined. This study investigates whether CNN feature extractors reliably encode front-back vowel distinctions, drawing parallels to hierarchical representations in image recognition \cite{yan2015hdcnnhierarchicaldeepconvolutional}. We address the following research questions:

RQ1: To what extent do convolutional layers in Wav2Vec encode phonetic distinctions, particularly for front-back vowel classification?

RQ2: How do CNN activations compare with MFCCs and formants in capturing front-back vowel differences?

\section{Materials and Method}
\subsection{Data}

The TIMIT corpus \cite{garofolo1993timit} is a widely used dataset for speech recognition, featuring 5.4 hours of American English speech from 630 speakers across eight dialects. Each speaker recited ten phonetically rich sentences, totaling 6,300 sentences with time-aligned phonetic, word, and orthographic transcriptions. The dataset includes 61 phonemes, with 17 vowels (monophthongs and diphthongs). Due to the imbalanced vowel distribution, we focus on nine monophthongs (i, I, E, {, A, O, o, U, u) for their clearer acoustic boundaries \footnote{The division of vowels into front and back is a purely analytical distinction, without implying any specific acoustic or articulatory status of \{}. This selection ensures coverage of both front and back vowels, aligning with our goal of probing CNN layers for front and back vowel classification. Table \ref{tab:vowel_durations} presents their mean duration and frequency.

\begin{table}[htbp]
    \centering
        \small
    \setlength{\tabcolsep}{3pt} 
    \renewcommand{\arraystretch}{0.9}
    \begin{tabular}{lrrl}
        \hline
        \textbf{Vowel} & \textbf{Mean Duration} & \textbf{Total Segments} & \textbf{Class} \\
        \hline
        i  & 1485.70 & 2359 & Front \\
        I  & 1250.49 & 2164 &  \\
        e  & 1502.39 & 1713 &  \\
        \{  & 2197.03 & 1085 &  \\
        \hline
        A  & 1954.88 & 999  & Back \\
        O  & 1944.24 & 871  &  \\
        o  & 2048.38 & 845  &  \\
        u  & 1540.61 & 837  &  \\
        U  & 1217.78 & 268  &  \\
        \hline
    \end{tabular}
    \caption{Mean Duration and Total Segments for Selected Vowels, with Front/Back Classification}
    \label{tab:vowel_durations}
    \vspace{-0.5cm}
\end{table}

\subsection{Data Preparation}
To ensure consistency, we included only vowels within the 1500-2000 sample range, reducing duration variability but potentially introducing bias by excluding shorter or longer vowels, which future work should address. This filtering resulted in 1,736 front and 946 back vowel segments. Since some segments were shorter than 2000 samples, we applied zero-padding to standardize all samples, ensuring consistency in feature extraction.

Figure \ref{fig:vowel_durations} (left) shows segment counts per vowel, while the right panel illustrates duration distribution post-filtering and padding. For acoustic features, we extracted 13 MFCCs using \texttt{Torchaudio} (Hamming window, 25ms window length, 10ms hop length) and F1/F2 formants with \texttt{parselmouth}. However, formants were not speaker-normalized, which may introduce variability—a limitation for future refinement.
\begin{figure}[htbp]
    \centering
    \includegraphics[width=\columnwidth]{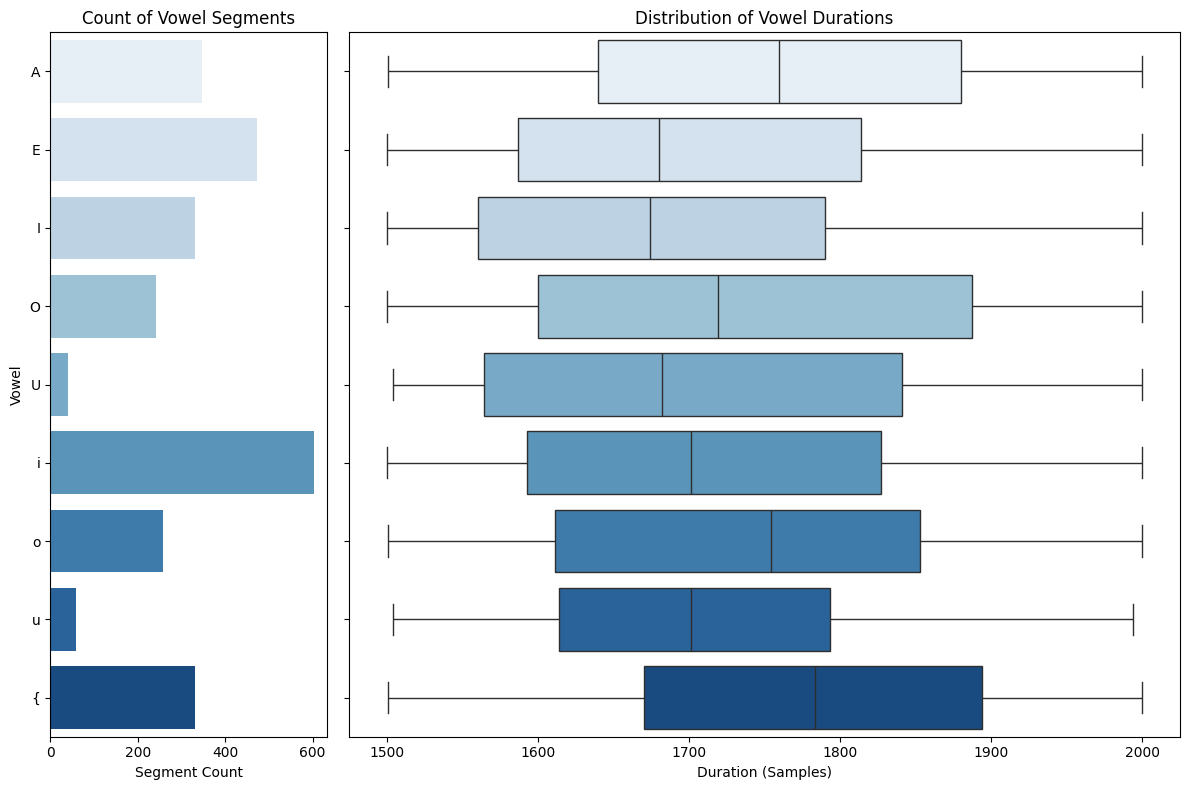}
    \vspace{-20pt}
    \caption{Vowel duration distribution after filtering and zero-padding.}
    \label{fig:vowel_durations}
\end{figure}

\subsection{Probing}

With the dataset standardized, we examined how Wav2Vec 2.0's convolutional layers encode phonetic information for front-back vowel classification. Using \texttt{facebook/wav2vec2-xlsr-53-espeak-cv-ft} \cite{xu2021simple}, a fine-tuned Wav2Vec 2.0 model optimized for phoneme recognition, we extracted inputs and outputs from its seven CNN layers. This model, pre-trained on 53 languages, retains Wav2Vec 2.0’s core architecture: a CNN feature extractor and a transformer network. The CNN, with seven layers, processes raw waveforms into structured acoustic embeddings, capturing local phonetic cues while reducing dimensionality. The transformer models long-range dependencies. We used the extracted CNN activations to train classifiers, evaluating their ability to distinguish front and back vowels.
\section{Experiments}

To evaluate the effectiveness of different feature representations for front and back vowel classification, we employed Support Vector Machine (SVM) classifiers, which are well-suited for high-dimensional data and effective in handling non-linearly separable classes through kernel methods. The feature representations used for classification can be grouped into two main categories: digital signal processing phonetic features, which explicitly encode linguistic knowledge and learned representations from deep models, which capture abstract, data-driven features.
The specific feature sets used in our experiments include: 13 Mel-Frequency Cepstral Coefficients (MFCCs) using \texttt{Torchaudio}, MFCCs combined with the first and second formant (F1 and F2) extracted using \texttt{parselmouth}, outputs of the feature extractor seven convolutional layers of the \texttt{wav2vec2-xlsr-53-espeak-cv-ft} model. While MFCCs were normalized using \texttt{MinMaxScaler}, ensuring values were scaled between 0 and 1 before training, CNN activations were used in their raw form. Since CNN activations vary in scale across layers, this may introduce differences in classifier performance that should be investigated in future work.

\subsection{Classifiers and Hyperparameter Search}

To ensure interpretability and a clear comparison of feature representations, we used Support Vector Machines (SVM) instead of neural networks for front-back vowel classification. SVMs offer well-defined decision boundaries, allowing controlled evaluation of feature discriminability. We performed binary (front-back) classification using SVMs, optimizing hyperparameters via Grid Search with 5-fold cross-validation. C values (0.5–5.0) were tested for regularization, and we compared \texttt{linear}, \texttt{polynomial}, and \texttt{RBF} kernels, expecting \texttt{RBF} to perform best for CNN activations. We tuned the gamma parameter (scale vs. auto) and tested one-vs-rest (\texttt{OVR}) and one-vs-one (\texttt{OVO}) decision strategies. A 300MB cache was allocated for efficiency, and CNN-based SVM models were trained separately for each convolutional layer.
\\section{Results}
\begin{figure}[!htbp]
    \centering
    \begin{minipage}{\linewidth}
        \centering
        \includegraphics[width=1\linewidth]{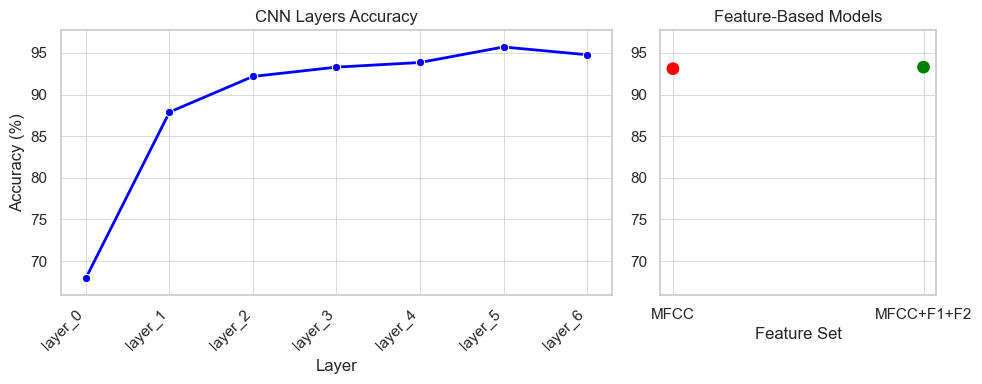}
        \vspace{-20pt}
        \caption{Accuracy trend}
        \label{fig:accuracy_trend}
    \end{minipage}
    
    \vspace{0.5cm}

    \begin{minipage}{\linewidth} 
        \centering
        \small
        \renewcommand{\arraystretch}{1.2}
        \setlength{\tabcolsep}{2pt}
        \begin{tabular}{l c c c c c c}
            \hline
            \textbf{Feature} & \textbf{C} & \textbf{Kernel} & \textbf{Gamma} & \textbf{Decision}  & \textbf{Accuracy} \\
            \hline
            Layer 0 & 5.0 & rbf & scale & ovr & 0.6797 \\
            Layer 1 & 4.0 & rbf & scale & ovr & 0.8790 \\
            Layer 2 & 3.0 & rbf & scale & ovr  & 0.9218 \\
            Layer 3 & 1.0 & poly & scale & ovr & 0.9330 \\
            Layer 4 & 1.5 & poly & scale & ovr & 0.9385 \\
            Layer 5 & 2.5 & rbf & scale & ovr & \textbf{0.9572} \\
            Layer 6 & 3.0 & rbf & scale & ovr & 0.9479 \\
            \hline
            MFCC & 0.5 & rbf & scale & ovr & 0.9311 \\
            MFCC + F1 + F2 & 3.5 & rbf & scale & ovr & 0.9330 \\
            \hline
        \end{tabular}
        \vspace{-9pt}
        \caption{Best hyperparameters and accuracy scores for SVM classifiers using different feature sets}
        \label{tab:svm_results}
    \end{minipage}
\end{figure}

We optimized SVM classifiers using GridSearchCV, training 1,080 models across different feature sets (MFCCs, MFCC+F1+F2, CNN activations) and hyperparameters. Accuracy, the primary metric, determined the best-performing models, reported in Table \ref{tab:svm_results}. CNN activations outperformed spectral features, probably benefiting from contextual information learned during pretraining. Accuracy improved from 67.97\% at layer 0 to 95.71\% at layer 5, with performance stabilizing in deeper layers (94.79\% at layer 6). The best CNN model used an \texttt{rbf} kernel with
C=2.5. MFCC-based models performed well, reaching 93.10\% with MFCCs alone and 93.29\% with F1/F2 added (C=3.5, \texttt{rbf}). While CNN activations from layer 5 achieved the highest accuracy, MFCC+F1+F2 was comparable to mid-level CNN layers, particularly layer 3.

\begin{figure*}[!htbp]
    \centering
    \includegraphics[width=0.32\textwidth]{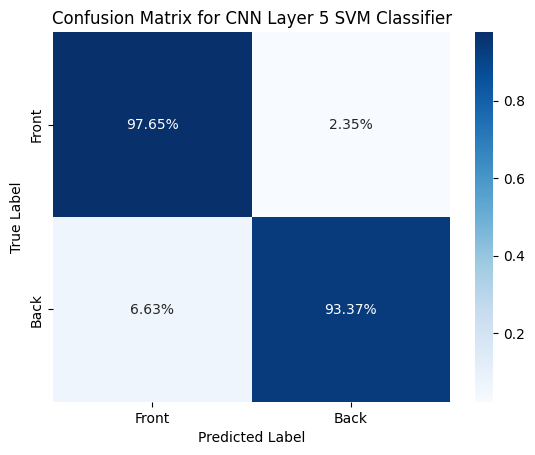}
    \includegraphics[width=0.32\textwidth]{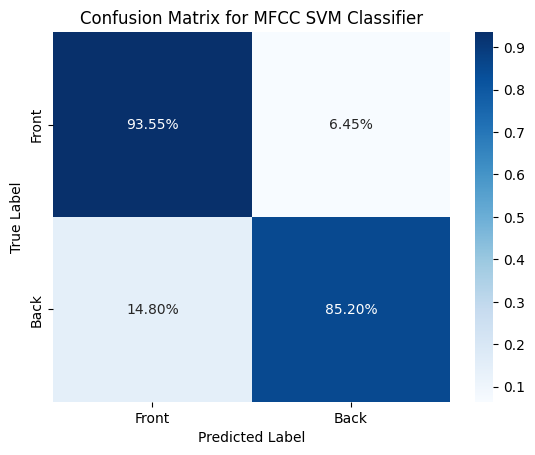}
    \includegraphics[width=0.30\textwidth]{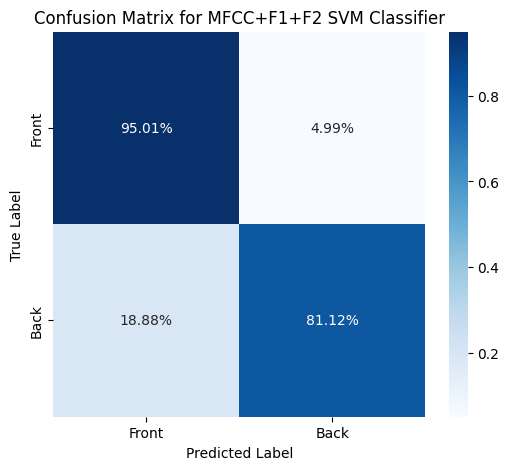}
    \vspace{-10pt}
    \caption{Comparison of confusion matrices for different classifiers}
    \label{fig:confusion_matrices}
\end{figure*}

\section{Discussion}
\begin{table}[!htbp]
    \centering
    \resizebox{\columnwidth}{!}{
    \begin{tabular}{cccccccc}
        \toprule
        \textbf{Layer} & 0 & 1 & 2 & 3 & 4 & 5 & 6 \\
        \midrule
        \textbf{MI} & 0.0819 & 0.1040 & 0.0934 & 0.0390 & 0.0522 & 0.0078 & 0.0257 \\
        \bottomrule
    \end{tabular}
    }
    \caption{Mutual Information between MFCCs and CNN activations at each layer.}
    \label{tab:mi_results}
\end{table}
The strong performance of MFCC-based classifiers suggests that spectral information is crucial for front vs. back vowel classification. Given that early CNN layers achieve comparable accuracy to MFCCs, it is likely that convolutional filters encode similar spectral features before progressively transforming them into more abstract representations. To better understand how MFCC-based spectral features relate to CNN activations, we computed Mutual Information regression with \texttt{scikit-learn} metrics (MI) between MFCCs and CNN activations at each convolutional layer \cite{pasad2022layerwiseanalysisselfsupervisedspeech}. We computed MI regression using 10 neighbors to ensure a balance between capturing local dependencies in the feature space and maintaining robustness against noise, allowing for a more reliable estimation of nonlinear relationships between input features and vowel classes. The results show that MI is highest in the early CNN layers, particularly at layer\_1 (0.1040) and layer\_2 (0.0934), aligning with strong classification performance (87.9\% and 92.1\% accuracy, respectively). This suggests that early CNN layers act as spectral feature extractors, similar to MFCCs. To ensure a fair comparison between different feature sets, we evaluated the classification performance of MFCCs, MFCC+F1+F2, and CNN activations using the same train-test split. The CNN-based classifier at layer\_5 achieved the highest accuracy (95.7\%), outperforming the MFCC-based models. However, the MFCC-only classifier still performed competitively (93.1\%), reinforcing the importance of spectral information in front-back vowel discrimination. The confusion matrices in Figure \ref{fig:confusion_matrices} highlight some key trends across models. All classifiers performed better on front vowels than back vowels. The CNN model exhibited fewer misclassifications overall, particularly in distinguishing back vowels. Interestingly, the MFCC-based classifiers misclassified more back vowels as front vowels.

\section{Conclusion}

This study investigated to which extent the convolutional layers in the Wav2Vec feature extractor capture phonetic information relevant to front-back vowel classification. Our results highlight that the CNN layers encode vowel-related features effectively, with classification accuracy steadily improving from the early to mid-level layers, peaking at layer\_5. This finding suggests that the convolutional feature extractor is not merely replicating traditional spectral features like MFCCs but transforming them into more abstract representations beneficial for ASR task. The comparison with MFCC-based classifiers indicates that while spectral information is crucial, deeper CNN layers refine vowel representations beyond what traditional handcrafted features capture. However, CNN representations are shaped by a model that has been exposed to more phonetic data, whereas MFCCs remain purely local representation of the signal. The greater abstraction observed in the CNN representations of Wav2Vec likely stems from the contextual representations learned during pre-training. In addressing RQ1, our analysis confirms that the CNN layers in Wav2Vec progressively encode phonetic information essential for front-back vowel classification, with deeper layers achieving higher accuracy. Regarding RQ2, mutual information analysis suggests that early CNN layers share similarities with MFCCs, but deeper layers transform these features into a more effective vowel representation. The highest classification accuracy obtained using CNN activations highlights the potential of self-supervised models to extract and refine phonetic information without reliance on predefined acoustic features. This aligns with Pasad et al. findings on the presence of phonetic information in the contextual layers \cite{pasad2022layerwiseanalysisselfsupervisedspeech}, and supports the idea that CNN encodes contextual information, influenced by the pre-training process. These observations undescore the remarkable effectiveness of MFCCs in contrast to CNN that required extensive pre-training with contextual layers on 60k hours of speech data to achieve a comparable level of performance, albeit constrained to this specific task.
This study aims to introduce a new perspective for probing the phonetic encoding capabilities of convolutional layers in self-supervised speech models, emphasizing the use of interpretable acoustic features. By comparing CNN intermediate layers output with traditional MFCC-based representations, we aimed to assess how vowel-related information is processed across different layers of the Wav2Vec feature extractor. Our findings indicate that while CNN layers progressively refine phonetic representations, traditional features like MFCCs and formant-based features provide a transparent and interpretable baseline for phonetic analysis. This highlights the potential of integrating such digital signal processing features when probing deep speech models, particularly for understanding how they encode linguistic information.

\section{Limitations}
The analysis is based on a small dataset compared to the extensive pre-training data of Wav2Vec, potentially limiting generalizability. Additionally, vowels were treated as isolated snapshots rather than as part of a continuous speech sequence, which may not fully capture dynamic phonetic variations. While CNN layers refine phonetic representations, they require extensive pre-training to match the effectiveness of MFCCs, which remain a transparent and interpretable baseline. Future work should explore how these representations generalize across phonetic contexts and different ASR architectures.
\section*{Acknowledgements}
Funded by the European Social Fund Plus Project code ESF2\_f3\_0003 “Excellence Scholarships for PhD students on topics of strategic relevance for South Tyrol”

\bibliography{anthology,custom}
\bibliographystyle{acl_natbib}

\end{document}